\pdfoutput=1

\documentclass[11pt]{article}

\usepackage[]{acl}

\usepackage{times}
\usepackage{latexsym}

\usepackage{graphicx}
\usepackage{multirow}
\usepackage{array}
\usepackage{makecell}

\usepackage[T1]{fontenc}

\usepackage[utf8]{inputenc}

\usepackage{microtype}

\title{Named Entity Inclusion in Abstractive Text Summarization}

\author{Sergey Berezin \\
  École des Mines de Nancy \\
   LORIA, UMR 7503, \\
   Université de Lorraine, 
   CNRS, Inria, \\ 54000 Nancy, France \\
  \texttt{sergeyberezin123@gmail.com} \\\And
  Tatiana Batura \\
  A. P. Ershov Institute of Informatics Systems \\
  630090 Novosibirsk, Russia \\
  \\
  \texttt{tatiana.v.batura@gmail.com} \\}

\begin{document}
\maketitle
\begin{abstract}
We address the named entity omission - the drawback of many current abstractive text summarizers. We suggest a custom pretraining objective to enhance the model's attention on the named entities in a text. At first, the named entity recognition model RoBERTa is trained to determine named entities in the text. After that, this model is used to mask named entities in the text and the BART model is trained to reconstruct them. Next, the BART  model is fine-tuned on the summarization task. Our experiments showed that this pretraining approach improves named entity inclusion precision and recall metrics.

\end{abstract}

\section{Introduction}
Current state-of-the-art abstractive summarization methods achieved significant progress, yet they are still prone to hallucinations and substitution of the named entities with vague synonyms or omitting mention of some of them at all \citep{4}, \citep{5}, \citep{6}. Such inconsistencies in the summary limit the practicability of abstractive models in real-world applications and carry a danger of misinformation. Example in Table \ref{tab:1} demonstrates the difference that named entity inclusion could make in the generated summary.

Scientific texts are especially vulnerable to this issue. Omitting or substituting the name of the metric used or the method applied can make a summary useless or, in the worst case scenario, totally misleading for a reader.


We make the following contributions:
\begin{itemize}
\item present a new method for pretraining a summarization model to include domain-specific named entities in the generated summary;

\item show that the BART model with the Masked Named Entity Language Model (MNELM) pretraining procedure is able to achieve higher precision and recall metrics of named entity inclusion.
\end{itemize}

\begin{table}[t]
\small
\centering
\begin{tabular}{ll}
\hline
\textbf{Without named entities} & \textbf{With named entities} \\
\hline
& \\
Famous North-American & \textbf{Andrew Ng} from \textbf{Stanford} \\
scientist suggested  & \ suggested a new way \\
a new way of training  & of training \textbf{feed-}\\
AI algorithms. &\textbf{forward neural networks}. \\
\hline
\end{tabular}
\caption{Example of NE omission}
\label{tab:1}
\end{table}

\section{Related work}

For automatic summarization, one of the important issues is extrinsic entity hallucinations, when some entities appear in summary, but do not occur in the source text \citep{maynez-etal-2020-faithfulness,DBLP:conf/naacl/PagnoniBT21}. A number of studies have been devoted to this problem, such as fixing entity-related errors \citep{nan-etal-2021-entity}, ensuring the factual consistency of generated summaries \citep{cao-etal-2020-factual}, and task-adaptive continued pertaining \citep{gururangan-etal-2020-dont}.
In our paper, we address the problem of named entity awareness of the summarization model by first training it on the NER task before final finetuning to make the model entity aware.

The idea of utilizing named entities during the pretraining phase first was described back in \citep{17}, 
where the authors proposed the usage of knowledge graphs by randomly masking some of the named entity alignments in the input text and asking the model to select the appropriate entities from the graphs to complete the alignments. One of the disadvantages of that approach is the need for a knowledge base, which is extremely difficult to build. Only a limited number of domain-specific knowledge bases exist, and none of them can be considered complete. 

The study \citep{18} addresses the problem of the factual consistency of a generated summary by a weakly-supervised, model-based approach for verifying factual consistency and identifying conflicts between source documents and a
generated summary. Training data is generated by applying a series of rule-based transformations to the sentences of the source documents. 

A similar approach is suggested by the authors of the paper 
\citep{19} who try to preserve the factual consistency of abstractive summarization by specifying tokens as constraints that must be present in the summary. They use a BERT-based keyphrase extractor model to determine the most
important spans in the text (akin to the extractive summarization) and then use these spans to constrain a generative algorithm. The big drawback of this approach is the vagueness of the keyphrases and the limited amount of training data. Also, the use of the BERT model leaves room for improvement.

The analogous solution uses 
\citep{20}, where the authors suggest entity-level content planning, i.e. prepending target summaries with entity chains – ordered sequences of entities that should be mentioned in the summary. But, as the entity chains are extracted from the reference summaries during the training, this approach cannot be used in an unsupervised manner, like MNELM, proposed in this work.

\section{Method}

We propose a three-step approach that aims to avoid all the aforementioned drawbacks:
1) at the first step the NER model is trained on a domain-specific dataset;
2) then the trained NER model is used for the MLM-like unsupervised pretraining of a language model;
3) the pretrained model is finetuned for the summarization task.

By following these steps, we can use a large amount of unlabeled data for the pretraining model to select domain-specific named entities and therefore to include
them in the generated summary. In comparison with a regular MLM pretraining, the suggested approach helps the model converge faster, shows an increased number of entities included in the generated summary, and drastically improves the avoiding of hallucinations, i.e. eliminates named entities that did not appear in the original text.

\section{Datasets and evaluation metrics}

In this work, we use two datasets: SCIERC \citep{21} for training named entity extraction model and ArXiv \citep{22} dataset for pretraining and training of the summarization
model. The SCIERC dataset includes annotations for scientific entities for 500 scientific abstracts. These abstracts are taken from 12 AI conference/workshop proceedings in four AI communities from the Semantic Scholar Corpus. These conferences include general AI (AAAI, IJCAI), NLP (ACL, EMNLP, IJCNLP), speech (ICASSP, Interspeech), machine learning (NIPS, ICML), and computer vision (CVPR, ICCV, ECCV) conferences. The dataset contains 8.089 named entities and defines six types for annotating scientific entities: Task, Method, Metric, Material, Other-Scientific-Term and Generic. SCIERC utilizes a greedy annotation approach for spans and always prefers the longer span whenever ambiguity occurs. Nested spans are allowed when a subspan has a relation/coreference link with another term outside the span.

The second dataset is the Arxiv dataset which takes scientific papers as an example of long documents and their abstracts are used as ground-truth summaries. Authors of the dataset removed the documents that are excessively long or too short, or do not have an abstract or some discourse structure. Figures and tables were removed using regular expressions to only preserve the textual information. Also, math formulae and citation markers were normalized with special tokens. Only the sections up to the conclusion section of the document were kept for every paper.

This dataset contains 215,912 scientific papers with the average length of 4,938 words and the average summary length of 220 words. To evaluate the performance of the model we used ROUGE-1, ROUGE-2, and ROUGE-L metrics.

For scoring the occurrence of named entities and their soundness and completeness we use named-entity-wise precision and recall:

\small
$$NE\;precision = \frac{correct \;NE\; in\; summary}{number\; of\; NE\; in\; summary}$$

$$NE\;recall = \frac{correct \;NE\; in\; summary}{number\; of\; NE\; in\; source}$$
\normalsize
\section{Experiments}
\begin{figure}[h]
\includegraphics[scale=0.5]{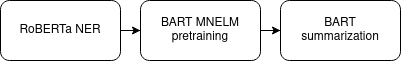}
\caption{Training sequence}
\label{fig:1}
\centering
\end{figure}

The training procedure of our model consists of the three main stages, illustrated in Figure \ref{fig:1}. 

\subsection{NER preparation}
To start our pipeline, we trained the Named Entity Recognition model. For this purpose, we used the RoBERTa \citep{25} language model. After the training for 7 epochs, we obtained an F1 macro score of 0.51 on the test dataset.

\subsection{Custom LM pretraining} 

BART \citep{27} uses the standard sequence-to-sequence Transformer architecture 
\citep{28} and it is pretrained by corrupting documents and then optimizing a reconstruction loss – the cross-entropy between the decoder’s output and the content of the original document. Unlike most of the existing denoising autoencoders, which are tailored to specific noising schemes, BART allows us to apply any type of document corruption. In the extreme case, where all information about the source is lost, BART is equivalent to a regular language model.

This unique ability opens the road to usage of our previously trained NER model. We use it to find named entities in scientific texts from the ArXiv dataset and substitute them with [mask] tokens. This way, we bring the model's attention to the named entities instead of just random words, most of which might be from a general domain. In our experiments, we used a 0.5 probability of masking.

This approach was inspired by the original BART paper, in the conclusion of which authors encourage further experiments with noising functions: “Future work should explore new methods for corrupting documents for pre-training, perhaps tailoring them to specific end tasks” \citep{27}.

We pretrained on 215,912 scientific articles on a single epoch starting with a learning rate of 5 * 10$^{-5}$ and a linear scheduler with gamma = 0.5 every 10,000 steps.

\subsection{Summarization training} 
After pretraining the BART model, we finetuned it on a summarization task. Because BART has an autoregressive decoder, it can be directly fine-tuned for sequence generation tasks such as abstractive question answering and summarization. In both of these tasks, information is copied from the input, but manipulated, which is closely related to the denoising pre-training objective. Here, we trained BART with a batch size of 1 for a single epoch. We figured out that the model easily overfits, so we had to use a learning rate scheduled every 5,000 steps with gamma = 0.5. The initial learning rate was set to be 2 * 10$^{-5}$. For training we used NVIDIA Tesla K80 GPU, the training took around 30 hours.

\section{Results}
Our model shows higher precision and recall in named entity inclusion in comparison to the same architecture, which was pretrained using regular masked language model objective - results of both models can be found in Table \ref{tab:2}. Examples of generated summaries are shown in Appendix \ref{sec:appendix}.

\begin{table}[t]
\centering
\begin{tabular}{lcc}
\hline
 & \textbf{MNELM} & \textbf{MLM}\\
\hline

NE Precision & \textbf{0.93} & 0.86\\
NE Recall & \textbf{0.39} & 0.38\\ 

\hline
\end{tabular}
\caption{Named Entity inclusion scores.}
\label{tab:2}
\end{table}

\begin{table}[t]
\centering
\begin{tabular}{llcc}
\hline
& & \textbf{MNELM} & \textbf{MLM}\\
\hline
\multirow{3}*{ROUGE-1} & F1 & \textbf{0.36} & 0.35\\
& precision & \textbf{0.51} & 0.49\\ 
& recall & 0.29 & 0.29\\
\hline
\multirow{3}*{ROUGE-2} & F1 & \textbf{0.13} & 0.12\\ 
& precision & \textbf{0.21} & 0.19\\
& recall & 0.10 & 0.10\\ 
\hline
\multirow{3}*{ROUGE-L} & F1 & \textbf{0.32} & 0.31\\
& precision & \textbf{0.45} & 0.43\\ 
& recall & \textbf{0.26} & 0.25\\
\hline
\end{tabular}
\caption{Summarization scores. MNELM was trained for 20k steps, MLM was trained for 25k steps.}
\label{tab:3}
\end{table}

\section{Discussion}
During the training of our model, we noticed that increase in common metrics for text summarization causes a decrease in named entity inclusion. We believe the reason for this is the limited length of the generated summary - one can have only so many named entities, before they will displace other words from the original text, causing the model to reformulate sentences and miss more words from the source. Therefore, during training, we tried to find the optimum point, at which the model will have high ROUGE scores and will still have high NE inclusion. At this point the MNELM-pretrained model, while keeping higher NE inclusion, converges faster than a regular MLM (in terms of ROUGE metrics). The comparison can be found in Table \ref{tab:3}. Obtained summarization scores are inferior to the recently published state of the art models like PRIMER \citep{29} (ROUGE-1 = 47.6; ROUGE-2 = 20.8) or DeepPyramidon \citep{30} (ROUGE-1 = 47.2; ROUGE-2 = 20), but their ability to preserve named entities in text is yet to be determined.


\section{Conclusion}
In this work, we described the task of preserving named entities in an automatically generated summary and presented the Masked Named Entity Language Model (MNELM) pretraining task. We show that with the MNELM pretraining procedure the BART model can achieve higher precision and recall of named entity inclusion. 

Pretraining with the MNELM task helps the model concentrate on domain-specific words, whereas MLM learns to reconstruct mostly common words. This leads to stronger attention on named entities, more likely preserving them in a generated text. The suggested model shows solid results in summarization metrics in comparison to the regular approach and converges faster.

In further research, we plan to improve the quality of the pretraining by masking a sequence of named entities with a single mask – the step that could help the model, according to the original BART paper \citep{27}. Also, we plan to conduct more experiments with different hyperparameters (such as masking probability), on more datasets, including PubMed \citep{22} and to train an even better NER model. In addition, we plan to improve the proposed model by overcoming the internal limitation on the number of input tokens (currently, it only has access to 1024 tokens).

\bibliography{anthology,custom}
\bibliographystyle{acl_natbib}

\appendix

\section{Appendix}
\label{sec:appendix}

Below is the comparison of the generated summaries. Named entities are in bold. First text is generated by the MNELM-pretrained model, second text is produced by the  MLM-pretrained model:\bigskip

1. "the problem of \textbf{admission control} for \textbf{web - based applications} is typically
considered as a problem of \textbf{system sizing} : enough resources are to be
provisioned to meet \textbf{quality of service} requirements under a wide range of
operating conditions. while this \textbf{approach} is beneficial in making the \textbf{site}
performance satisfactory in the most common working situations, it still leaves
the site incapable to face sudden and unexpected surges of traffic. in this
context , it is impossible to predict the intensity of the \textbf{overload}. this work is
motivated by the need to formulate a fast \textbf{reactive and autonomous approach}
to \textbf{admission control}. in particular, we propose an original \textbf{self- * overload
control policy} ( soc ) which enables some fundamental self - * properties such
as \textbf{self - configuration, self - optimization, self - protection}."\bigskip

2. "we propose an autonomous \textbf{approach} to \textbf{admission control} in \textbf{distributed
web systems}. the proposed \textbf{policy} is based on \textbf{self - configuration, self -
optimization,} and \textbf{self - protection}. in particular, the proposed \textbf{system} is
capable of self - configuring its \textbf{component level parameters} according to
performance requirements, while at the same time it optimizes its own
responsiveness to \textbf{overload}. at \textbf{session granularity} , it does not require any
prior knowledge on the incoming traffic and can be applied to \textbf{non - session
based} traffic as well."\bigskip

MNELM model scores: NE precision = 0.91; NE recall = 0.49.
MLM model scores: NE precision = 0.71; NE recall = 0.20.

\end{document}